\documentclass[letterpaper, 10 pt, conference]{ieeeconf}  

\IEEEoverridecommandlockouts                              

\usepackage{cite}
\usepackage{amsmath,amssymb,amsfonts}
\usepackage{xcolor}
\usepackage{ragged2e}
\usepackage{float}
\usepackage{algorithm}
\usepackage{algpseudocode}
\usepackage{amsmath,amssymb}
\usepackage{graphicx}
\usepackage{textcomp}
\usepackage{xcolor}
\usepackage{amsmath,amssymb,amsfonts}
\usepackage{mathtools}
\usepackage{graphicx}
\usepackage{textcomp}
\usepackage{xcolor}
\usepackage{tikz}
\usepackage{ragged2e}
\usepackage{units}
\usepackage{cprotect}
\usepackage{amsfonts}
\usepackage{fvextra}
\usepackage{textcomp}
\usepackage{paracol}
\usepackage{amssymb}
\usepackage{cite}
\usepackage{amsmath,amssymb,amsfonts}
\usepackage{mathtools}
\usepackage{amsmath}
\usepackage{amsmath}
\usepackage{graphicx}
\usepackage{booktabs}
\usepackage{soul}
\usepackage{graphicx}
\usepackage{textcomp}
\usepackage{xcolor}
\usepackage{tikz}
\usepackage{units}
\usepackage{tikz}
\usepackage{subcaption}
\usepackage{bm}
\usepackage{hyperref}
\usepackage{makecell}
\usepackage [english]{babel}
\usepackage [autostyle, english = american]{csquotes}
\usepackage{lipsum,graphicx,multicol,subcaption}
\usepackage [english]{babel}
\usepackage[justification=centering]{caption}
\usepackage{lipsum,graphicx,multicol}
\usepackage{atbegshi}

\usepackage{placeins}

\title{\LARGE \bf
When LiDAR Fades, Radar Takes Over:
LiDAR Reliability-Aware Adaptive Localization Under Airborne Obscurants
}
\author{
Moumita Mukherjee, Avijit Banerjee, Anton Koval, and George Nikolakopoulos%
\thanks{This work has been partially funded by the European Union's Horizon Europe
Research and Innovation Programme under Grant Agreement No.~101138451 PERSEPHONE,
and in part by the Swedish Department of Energy and LKAB through the Sustainable
Underground Mining (SUM) Academy Programme project ``Autonomous Drones for
Underground Mining Operations.''}%
\thanks{The authors are with the Robotics and AI Group, Department of Computer,
Electrical and Space Engineering, Lule\aa{} University of Technology,
Lule\aa{}, Sweden.}%
\thanks{Corresponding author: Moumita Mukherjee
(email: \texttt{moumuk@ltu.se}).}%
}
\begin{document}
\maketitle
\thispagestyle{empty}
\pagestyle{empty}
\begin{abstract}
LiDAR-inertial odometry (LIO) is widely used for real-time localization and mapping, but its reliability can degrade substantially in subterranean environments affected by smoke, dust, and other airborne obscurants, whereas radar remains comparatively robust. This article presents a one-class temporal learning framework for LiDAR degradation detection and reliability-aware localization under such conditions. A broad set of degradation-sensitive cues, including spatial, cross-modal, geometric/motion, intensity, and measurement-recency features, is evaluated to characterize changes in LiDAR sensing quality. Their temporal evolution is modelled over sliding windows using an LSTM autoencoder trained exclusively on nominal operation, enabling deviations from healthy sensor behaviour to be detected without requiring labelled degradation data. The resulting LiDAR reliability estimate governs adaptive selection between LIO and radar inertial odometry (RIO), retaining LIO during reliable sensing and transferring localization authority to RIO when degradation is detected. The framework is evaluated using real-world ground robot data collected in GNSS-denied subterranean tunnels under nominal and dense-smoke conditions. Experimental results demonstrate degradation detection, adaptive modality switching, and continued localization output under severe LiDAR degradation. A feature family ablation study further shows that a reduced feature representation can improve degradation-aware localization performance in the evaluated dense-smoke scenarios, highlighting the importance of feature parsimony in one-class sensor degradation detection. The code will be released as open source to support reproducibility and further research.
\end{abstract}
\section{Introduction}

The rapid advancement of LiDAR technology has significantly improved localization, mapping, and three-dimensional perception in challenging environments. However, in GNSS-denied subterranean environments such as tunnels and underground facilities, dense smoke, rain, and other airborne particulates can substantially degrade LiDAR measurements~\cite{zhang2021lidar,phillips2017dust}.
\begin{figure}[h]
\centering
\includegraphics[width=0.49\textwidth]{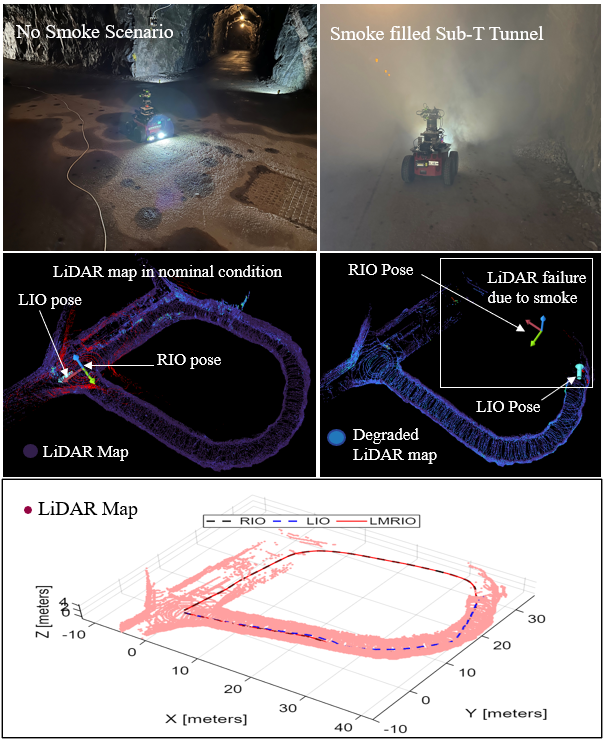}
\caption{Conceptual framework for LiDAR degradation-aware adaptive
localization under airborne obscurants.}
\label{fig:Cover_fig_1}
\end{figure}
Such degradation may reduce the number of valid returns, alter their spatial distribution, and change pointcloud density, thereby reducing the reliability of LiDAR-based perception in real-world deployments
~\cite{park2025lidar,vargas2020weather}. Although radar is comparatively robust under degraded visibility, LiDAR remains valuable for its high-resolution geometric information. Reliable localization therefore
requires both complementary sensing modalities and the ability to identify when LiDAR measurements become unreliable, as illustrated in Fig.~\ref{fig:Cover_fig_1}.

This motivates degradation-sensitive LiDAR feature extraction that captures changes in spatial occupancy, density, temporal consistency, and related sensing characteristics. LiDAR
degradation may be observed as abrupt changes, missing observations, or persistent deviations from nominal behaviour. Existing anomaly detection methods include distribution, clustering, distance, and density based approaches~\cite{jain1999data,knorr1997unified,knorr2000distance}.
However, these methods typically operate on individual observations and do not explicitly model the temporal evolution of sensor degradation. This limitation motivates the sequence based formulation adopted in this article.
\subsection{Feature Extraction for Anomaly Detection}

LiDAR pointclouds are unordered and variable in size, making direct temporal modelling computationally demanding. Representations based on subsampling, projection, or learned point-wise features can reduce this complexity, but may suppress local variations in point concentration and spatial occupancy that are informative for sensor degradation~\cite{qi2017pointnet,wu2023csi}. Voxel-based representations provide a structured alternative by preserving three-dimensional spatial organization while enabling compact characterization of return density and occupancy~\cite{zhao2025pvlf,lyu2024dynamic}. These statistics provide interpretable reliability cues through changes in the number, concentration, and spatial distribution of valid LiDAR returns. Accordingly, the proposed framework combines voxel-based spatial statistics with complementary temporal, geometric, cross-modal, inertial, and intensity related features for sequence based degradation detection.

\subsection{One-Class Temporal Degradation Detection with LSTM Autoencoders}
Conventional anomaly detection methods, including clustering, Isolation Forest, and support vector machines, can identify observations that deviate from nominal feature distributions~\cite{samariya2023comprehensive}. However, such approaches typically assess individual feature vectors and do not explicitly model the temporal evolution of sensor degradation. This is important for LiDAR reliability monitoring, where degradation may develop progressively over consecutive observations rather than as an isolated outlier. Autoencoder based methods provide a suitable one-class~\cite{khan2014one} learning framework by modelling nominal sensor behaviour without requiring labelled failure data. Deviations from the learned nominal representation are reflected by increased reconstruction error. To capture temporal dependencies, the proposed approach employs an LSTM autoencoder~\cite{ashraf2020novel} operating on sliding windows of multivariate degradation-sensitive features. By modelling the temporal evolution of nominal LiDAR measurements, the network identifies abnormal temporal patterns associated with emerging degradation.

Thus, the main contributions of this article are summarized as follows:

\begin{itemize}

\item A compact and interpretable multi-cue LiDAR reliability representation is developed using voxel based spatial statistics, measurement recency, LIO-RIO pose consistency, point-cloud geometry, inertial motion, and LiDAR intensity features. A feature family ablation study quantifies the contribution of these cues under dense smoke conditions.

\item A one-class temporal LiDAR degradation detector based on an LSTM autoencoder is proposed to model nominal sensor behaviour over sliding windows and identify emerging degradation from reconstruction error deviations without requiring labelled failure data.

\item A reliability aware adaptive LIO-RIO localization framework is introduced, in which LiDAR health governs online estimator selection. LIO is retained during reliable sensing, while RIO is selected when LiDAR degrades or becomes unavailable. The framework is evaluated using real-world data from GNSS-denied tunnel experiments under nominal and dense smoke conditions.

\end{itemize}

The remainder of this article is organized as follows. Sections~II and~III present the problem formulation and the proposed LMRIO framework, respectively. Section~IV presents the experimental evaluation, and Section~V concludes the article.
\subsection{Notation}
Let $k\in\mathbb{N}$ denote the discrete-time index on the common IMU timeline with timestamp $t_k$, and let $\eta\in\mathbb{N}$ index unique LiDAR scans. The robot's planar position is $\mathbf{p}_{k,xy}=[x_k,y_k]^\top$. The $\eta$th LiDAR scan is ${}^{\mathcal L}\mathcal{P}_\eta=\{\mathbf{p}_{\eta,i}\in\mathbb{R}^3\}_{i=1}^{N_\eta^{\mathcal L}}$, where $N_\eta^{\mathcal L}$ is the number of LiDAR returns. The corresponding twelve-dimensional reliability feature vector is $\mathbf{f}_\eta\in\mathbb{R}^{12}$, with standardized form $\bar{\mathbf{f}}_\eta$. A temporal window of $W$ consecutive feature vectors is denoted by $\mathbf{F}_\eta$, with LSTM-autoencoder reconstruction $\hat{\mathbf{F}}_\eta$, reconstruction error $e_\eta$, and anomaly threshold $\tau_{\mathrm{AE}}$. The final LiDAR degradation state on the IMU timeline is $\gamma_k\in\{0,1\}$, where $0$ and $1$ denote reliable and degraded/unavailable LiDAR, respectively. The LIO, RIO, and LMRIO planar position estimates are denoted by $\hat{\mathbf{p}}_{k,xy}^{\mathrm{LIO}}$, $\hat{\mathbf{p}}_{k,xy}^{\mathrm{RIO}}$, and $\hat{\mathbf{p}}_{k,xy}^{\mathrm{LMRIO}}$, respectively. The operators $\|\cdot\|_2$, $\|\cdot\|_F$, and $\lfloor\cdot\rfloor$ denote the Euclidean norm, Frobenius norm, and component-wise floor operator. 
\section{Problem Formulation}
\label{sec:problem formulation}
This article considers a non-holonomic ground robot operating in a GNSS-denied subterranean environment and equipped with radars, a 3D LiDAR, and an IMU. Under nominal conditions, LiDAR provides reliable geometric information for LiDAR inertial odometry (LIO)~\cite{xu2022fast}. Under airborne obscurants such as smoke and dust, however, LiDAR measurements may progressively degrade through reduced or corrupted returns, altered spatial density and intensity, and measurement staleness. Since these changes may also arise from scene geometry, surface properties, or robot motion, LiDAR degradation cannot be robustly characterized by a single measurement cue~\cite{phillips2017dust}. The objective is therefore to estimate LiDAR reliability from complementary sensor and odometry cues, detect departures from nominal LiDAR behaviour, and determine when LiDAR-based localization becomes unreliable. The resulting degradation estimate is used to adaptively select between LIO during reliable LiDAR operation and radar inertial odometry (RIO) when LiDAR is degraded or unavailable, thereby maintaining localization in GNSS-denied obscurant conditions.
\section{Methodology}
\label{sec:methodology}
\begin{figure*}[t]
    \centering
    \includegraphics[width=0.99\textwidth]
    {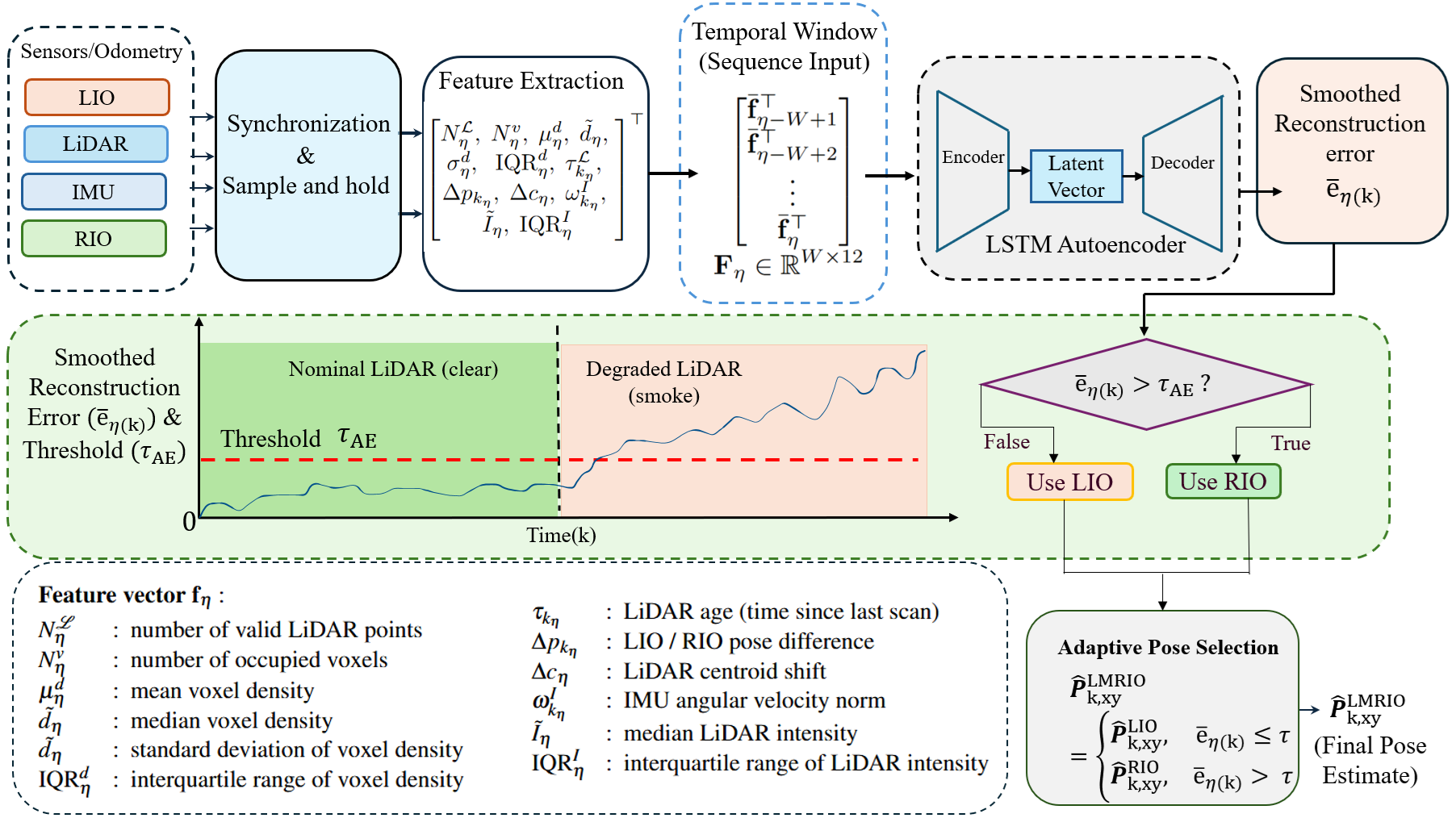}
    \caption{\justifying Proposed reliability aware LMRIO framework. Asynchronous LiDAR, IMU, LIO, and RIO measurements are temporally synchronized and encoded into a Twelve dimensional degradation sensitive feature representation($f_\eta$). Consecutive normalized feature vectors($\bar{f}_\eta$) form a temporal sequence processed by an LSTM autoencoder trained on nominal operation. Reconstruction errors are used to infer LiDAR degradation and adaptively select between LIO and RIO.}
    \label{fig:Framwork_LMRIO}
\end{figure*}

The proposed LiDAR Multi-Radar Inertial Odometry (LMRIO) framework performs reliability-aware localization by adaptively selecting between LIO and RIO based on LiDAR health in dense smoke and clear-air nominal conditions. As shown in Fig.~\ref{fig:Framwork_LMRIO}, asynchronous LiDAR, IMU, LIO, and RIO measurements are synchronized, and each LiDAR scan is encoded by a twelve-dimensional degradation-sensitive feature vector. Consecutive features are processed by an LSTM autoencoder trained on nominal data, while reconstruction error determines the degradation state and triggers switching to RIO when LiDAR becomes unreliable.
\subsection{Sensor Synchronization}
\label{sec:synchronization}
LiDAR, IMU, LIO, and RIO measurements are acquired asynchronously at different sampling rates. The IMU timestamps define the common reference timeline, with $t_k$ denoting the $k$th IMU timestamp. Let
$\mathcal{S}=\{\mathcal{L},\mathrm{LIO},\mathrm{RIO}\}$ denote the lower-rate sensing and odometry streams. For each modality $s\in\mathcal{S}$, let
$\{t_i^s,\mathbf{z}_i^s\}_{i=1}^{N_s}$ denote its $N_s$ time-stamped measurements. At time $t_k$, the most recent available measurement is selected as
\begin{equation}
\kappa_s(k)=
\max\left\{
i \;\middle|\; t_i^s \leq t_k
\right\},
\label{eq:sync_index}
\end{equation}
and synchronized to the IMU timeline via $\tilde{\mathbf{z}}_k^s
=\mathbf{z}_{\kappa_s(k)}^s$. For LiDAR, the most recent scan index available at $t_k$ is
\begin{equation}
\eta(k)\triangleq\kappa_{\mathcal{L}}(k),
\label{eq:lidar_index_mapping}
\end{equation}
such that
\begin{equation}
\tilde{\mathbf{z}}_k^{\mathcal{L}}
=
{}^{\mathcal{L}}\mathcal{P}_{\eta(k)}.
\label{eq:lidar_sync}
\end{equation}
This zero-order hold provides temporally aligned LiDAR, LIO, and RIO measurements on the common IMU timeline. Repeated LiDAR observations introduced by the hold operation are used only for synchronization and are excluded from the LSTM input sequence. Hence, temporal sequences are constructed solely from unique LiDAR scans indexed by $\eta$.
\subsection{Voxel-Grid based LiDAR Pointcloud Representation}
\label{sec:voxel_representation}
For each newly received LiDAR scan, the point cloud is represented as
\begin{align}
{}^{\mathcal L}\mathcal{P}_\eta
=
\left\{
\mathbf{p}_{\eta,i}\in\mathbb{R}^{3}
\right\}_{i=1}^{N_\eta^{\mathcal L}},
\qquad
\mathbf{p}_{\eta,i}
=
[x_{\eta,i},y_{\eta,i},z_{\eta,i}]^\top
\end{align}
where $N_\eta^{\mathcal L}$ denotes the number of LiDAR returns per scan. To obtain a compact representation of the spatial distribution of the point cloud, ${}^{\mathcal L}\mathcal{P}_\eta$ is partitioned into cubic voxels of side length $v$. Defining the minimum pointcloud
coordinate as
\begin{equation}
\mathbf{p}^{\min}_\eta
=
\min_{\mathbf{p}_{\eta,i}\in{}^{\mathcal L}\mathcal{P}_\eta}
\mathbf{p}_{\eta,i},
\end{equation}
the voxel index associated with point $\mathbf{p}_{\eta,i}$ is
\begin{equation}
\mathbf{q}_{\eta,i}
=
\left\lfloor
\frac{
\mathbf{p}_{\eta,i}-\mathbf{p}^{\min}_\eta
}{v}
\right\rfloor .
\label{eq:voxel_index}
\end{equation}
Let $\mathcal{V}_\eta$ denote the set of unique occupied voxels and $N_\eta^v=|\mathcal{V}_\eta|$ their number. If $n_{\eta,j}$ points occupy voxel $j$, its point density is defined as
\begin{equation}
d_{\eta,j}
=
\frac{n_{\eta,j}}{v^3},
\qquad
j=1,\ldots,N_\eta^v.
\label{eq:voxel_density}
\end{equation}
The occupied voxel density distribution
\begin{equation}
\mathcal{D}_\eta
=
\{d_{\eta,1},\ldots,d_{\eta,N_\eta^v}\}
\end{equation}
is summarized by its mean $\mu_\eta^d$, median $\tilde{d}_\eta$,
standard deviation $\sigma_\eta^d$, and interquartile range
$\mathrm{IQR}_\eta^d$. Together with $N_\eta^{\mathcal L}$ and
$N_\eta^v$, these statistics characterize changes in both the amount and spatial distribution of LiDAR information.
\subsection{Degradation Sensitive Feature Extraction}
\label{sec:feature_extraction}
Voxel density statistics alone may not fully distinguish sensor degradation from changes induced by environment geometry or robot motion. The spatial representation is therefore augmented with temporal, cross-modal, geometric, inertial, and intensity-based cues.

Let
\begin{equation}
k_\eta
\triangleq
\min\left\{
k \mid \eta(k)=\eta
\right\}
\end{equation}
denotes the first IMU index at which the $\eta$th LiDAR scan is available. The age of the latest LiDAR observation on the IMU timeline is defined as
\begin{equation}
\tau_k^{\mathcal L}
=
t_k-t_{\kappa_{\mathcal L}(k)}^{\mathcal L},
\label{eq:lidar_age}
\end{equation}
where a larger $\tau_k^{\mathcal L}$ indicates an increasingly stale LiDAR measurement. Cross-modal localization consistency is characterized through the planar discrepancy between the synchronized LIO and RIO estimates,
\begin{equation}
\Delta p_k
=
\left\|
\hat{\mathbf{p}}_{k,xy}^{\mathrm{LIO}}
-
\hat{\mathbf{p}}_{k,xy}^{\mathrm{RIO}}
\right\|_2 .
\label{eq:pose_diff}
\end{equation}

The LiDAR pointcloud centroid is
\begin{equation}
\mathbf{c}_\eta
=
\frac{1}{N_\eta^{\mathcal L}}
\sum_{i=1}^{N_\eta^{\mathcal L}}
\mathbf{p}_{\eta,i},
\end{equation}
and its displacement between successive LiDAR observations is
\begin{equation}
\Delta c_\eta
=
\left\|
\mathbf{c}_\eta-\mathbf{c}_{\eta-1}
\right\|_2 .
\label{eq:centroid_shift}
\end{equation}

To provide rotational motion context, the magnitude of the measured IMU angular velocity is included as
\begin{equation}
\omega_k^I
=
\left\|
\boldsymbol{\omega}_k^I
\right\|_2 .
\label{eq:imu_ang_norm}
\end{equation}

Finally, let $\mathcal{I}_\eta$ denote the set of valid LiDAR return intensities associated with scan $\eta$. Its median and interquartile range are given by
\begin{equation}
\tilde{I}_\eta
=
\mathrm{median}(\mathcal{I}_\eta),
\qquad
\mathrm{IQR}_\eta^I
=
Q_{0.75}(\mathcal{I}_\eta)
-
Q_{0.25}(\mathcal{I}_\eta).
\label{eq:intensity_features}
\end{equation}

The resulting twelve-dimensional reliability feature vector associated with the $\eta$th LiDAR scan is
\begin{equation}
\mathbf{f}_\eta =
\begin{bmatrix}
N_\eta^{\mathcal L},\;
N_\eta^v,\;
\mu_\eta^d,\;
\tilde{d}_\eta,\;
\tilde{d}_\eta,\;
\mathrm{IQR}_\eta^d,\;\\
\tau_{k_\eta}^{\mathcal L},\;
\Delta p_{k_\eta},\;
\Delta c_\eta,\;
\omega_{k_\eta}^{I},\;
\tilde{I}_\eta,\;
\mathrm{IQR}_\eta^I
\end{bmatrix}^{\top}
\in\mathbb{R}^{12}.
\label{eq:feature_vector}
\end{equation}
\subsection{Feature Standardization and Temporal Sequence Construction}
Because the reliability features have different physical units and numerical ranges, each feature is standardized using statistics computed exclusively from nominal training data:
\begin{equation}
\bar{\mathbf{f}}_\eta
= (\mathbf{f}_\eta-\boldsymbol{\mu}_H)
\oslash
\boldsymbol{\sigma}_H,
\label{eq:normalization}
\end{equation}

where $\boldsymbol{\mu}_H$ and $\boldsymbol{\sigma}_H$ denote the feature-wise mean and standard deviation of the healthy training dataset, respectively, and $\oslash$ denotes element-wise division. A sliding temporal window containing $W$ consecutive unique LiDAR observations is then constructed as
\begin{equation}
\mathbf{F}_\eta
=
\begin{bmatrix}
\bar{\mathbf{f}}_{\eta-W+1}^{\top}\\
\bar{\mathbf{f}}_{\eta-W+2}^{\top}\\
\vdots\\
\bar{\mathbf{f}}_{\eta}^{\top}
\end{bmatrix}
\in\mathbb{R}^{W\times 12}.
\label{eq:temporal_window}
\end{equation}
Repeated LiDAR observations produced by zero-order hold are removed before sequence construction so that the temporal model is trained using unique sensor observations.
\subsection{One-Class Temporal Degradation Detection}
\label{sec:lstm_detection}

The temporal model follows a one-class learning strategy and is trained exclusively on nominal LiDAR operation. Let $g_{\theta}(\cdot)$ denote the LSTM autoencoder parameterized by
$\theta$. For an input window $\mathbf{F}_\eta$, the reconstructed
sequence is
\begin{equation}
\hat{\mathbf{F}}_\eta
=
g_{\theta}(\mathbf{F}_\eta).
\end{equation}

The model parameters are learned by minimizing the mean-squared
reconstruction error over the healthy training sequences. During
online operation, the reconstruction error for a complete temporal window is evaluated as
\begin{equation}
e_\eta
=
\frac{1}{12W}
\left\|
\mathbf{F}_\eta-\hat{\mathbf{F}}_\eta
\right\|_F^2.
\label{eq:reconstruction_error}
\end{equation}

The anomaly threshold is determined from the upper tail of the healthy reconstruction-error distribution,
\begin{equation}
\tau_{\mathrm{AE}}
=
Q_{0.99}
\left(
\{e_\eta^H\}
\right),
\label{eq:ae_threshold}
\end{equation}
where $e_\eta^H$ denotes the reconstruction errors obtained from nominal training windows. To suppress isolated reconstruction-error fluctuations, the online score is smoothed over the most recent $S$ estimates,
\begin{equation}
\bar{e}_\eta
=
\frac{1}{S_\eta}
\sum_{j=0}^{S_\eta-1}e_{\eta-j},
\qquad
S_\eta=\min(S,N_\eta^e),
\label{eq:error_smoothing}
\end{equation}
where $N_\eta^e$ denotes the number of reconstruction-error
estimates available up to LiDAR scan $\eta$ and the learned anomaly indicator on the IMU timeline is
\begin{equation}
\gamma_k^{\mathrm{AE}}
=
\mathbb{I}
\left(
\bar{e}_{\eta(k)}>\tau_{\mathrm{AE}}
\right).
\label{eq:ae_flag}
\end{equation}

\subsection{Reliability-Aware LIO-RIO Selection}
\label{sec:adaptive_switching}
Because the LSTM autoencoder is evaluated only when a new LiDAR scan is received, reconstruction error alone cannot detect complete measurement stalls. A deterministic LiDAR recency is therefore evaluated continuously on the IMU timeline:
\begin{equation}
\gamma_k^{F}
=
\mathbb{I}
\left(
\tau_k^{\mathcal{L}}>\tau_a
\right),
\label{eq:freshness_flag}
\end{equation}
where $\tau_a$ denotes the maximum admissible LiDAR measurement age. In the current implementation, $\tau_a=1.5$~s. The final degradation state combines the learned anomaly indicator and the LiDAR age criterion
\begin{equation}
\gamma_k
=
\gamma_k^{\mathrm{AE}}
\lor
\gamma_k^{F},
\label{eq:final_flag}
\end{equation}
where $\gamma_k=0$ denotes reliable LiDAR operation and $\gamma_k=1$ denotes degraded or unavailable LiDAR sensing. The resulting reliability decision governs the active localization modality,
\begin{equation}
\hat{\mathbf{p}}_{k,xy}^{\mathrm{LMRIO}}
=
\begin{cases}
\hat{\mathbf{p}}_{k,xy}^{\mathrm{LIO}},
& \gamma_k=0,\\[1mm]
\hat{\mathbf{p}}_{k,xy}^{\mathrm{RIO}},
& \gamma_k=1.
\end{cases}
\label{eq:lmrio_switch}
\end{equation}
Thus, LIO is retained while LiDAR remains reliable, whereas RIO provides the localization estimate when either anomalous LiDAR behaviour or complete measurement staleness is detected.
\begin{algorithm}[h]
\caption{Reliability-Aware LIO-RIO Localization}
\label{alg:lmrio}
\begin{algorithmic}[1]

\Require $\mathbf f_\eta\in\mathbb{R}^{12}$, $g_\theta$,
$(\boldsymbol{\mu}_H,\boldsymbol{\sigma}_H)$,
$W,S,\tau_{\mathrm{AE}},\tau_a$

\Ensure $\hat{\mathbf p}_{k,xy}^{\mathrm{LMRIO}}$

\For{$k=1,2,\ldots$}

    \If{new LiDAR scan}

        \State $\eta\gets\eta(k)$

        \State $\bar{\mathbf f}_\eta\gets
        (\mathbf f_\eta-\boldsymbol{\mu}_H)
        \oslash\boldsymbol{\sigma}_H$

        \State Construct $\mathbf F_\eta$
        according to~\eqref{eq:temporal_window}

        \State $\hat{\mathbf F}_\eta
        \gets g_\theta(\mathbf F_\eta)$

        \State $e_\eta\gets
        \frac{1}{12W}
        \|\mathbf F_\eta-\hat{\mathbf F}_\eta\|_F^2$

        \State $S_\eta\gets
        \min(S,N_\eta^e)$

        \State $\bar e_\eta\gets
        \frac{1}{S_\eta}
        \sum_{j=0}^{S_\eta-1}e_{\eta-j}$

        \State $\gamma_k^{\mathrm{AE}}\gets
        \mathbb I(\bar e_\eta>\tau_{\mathrm{AE}})$

    \Else

        \State $\gamma_k^{\mathrm{AE}}
        \gets\gamma_{k-1}^{\mathrm{AE}}$

    \EndIf

    \State $\gamma_k\gets
    \gamma_k^{\mathrm{AE}}
    \lor
    \mathbb I(\tau_k^{\mathcal L}>\tau_a)$

    \State $\hat{\mathbf p}_{k,xy}^{\mathrm{LMRIO}}
    \gets
    (1-\gamma_k)\hat{\mathbf p}_{k,xy}^{\mathrm{LIO}}
    +
    \gamma_k\hat{\mathbf p}_{k,xy}^{\mathrm{RIO}}$

\EndFor

\end{algorithmic}
\end{algorithm}
\section{Experimental Results}
\label{sec:evaluation}

The proposed LMRIO framework(Fig.~\ref{fig:Framwork_LMRIO}) was evaluated using a ground robot in a GNSS-denied real-world subterranean tunnel under nominal and dense-smoke conditions. The sensing platform comprised a 3D LiDAR, 
radars, and an IMU, providing LIO and RIO estimates. The IMU, LiDAR, LIO, and RIO streams operated asynchronously at approximately 200, 5, 97, and 58~Hz, respectively, and were synchronized on the common IMU timeline as described in Sec.~\ref{sec:synchronization}. Sensor specifications are provided in Tables~\ref{Table: 3D velodyne Lidar} and~\ref{Table:radar_specifications}. Four traversals were considered, comprising P-shaped and round-tunnel trajectories under both nominal and dense-smoke conditions. Artificial smoke was introduced to induce attenuation and backscattering of LiDAR returns and to evaluate the response of the proposed reliability
mechanism to progressive degradation and measurement loss. Since independent trajectory ground truth was unavailable in the subterranean environment, LIO is used as the comparison reference under nominal
conditions, whereas RIO provides the comparison reference under dense smoke. Consequently, the reported trajectory metrics characterize agreement between localization estimates rather than absolute localization accuracy.
\subsection{Experimental Protocol and Model Configuration}
\label{sec:experimental_protocol}

The one-class LSTM autoencoder is trained exclusively on nominal P-shaped traversal data using the twelve-feature reliability representation. Repeated LiDAR observations introduced by sample and hold synchronization are removed, and the remaining healthy features are standardized and arranged into windows of $W=100$ with unique LiDAR scans. The network uses 32 hidden units and is trained with Adam for 300 epochs, a mini-batch size of 32, and an initial learning rate of $10^{-3}$. The anomaly threshold is set to the 99th percentile of the healthy reconstruction error distribution, with a voxel size of $v=3$~m. During online evaluation, reconstruction error is updated only when a new LiDAR scan arrives and is smoothed over the most recent 50 estimates. LiDAR age is evaluated continuously on the IMU timeline, with $\tau_a=1.5$~s used to detect stale measurements. 
\begin{figure*}[h]
    \centering
    \includegraphics[width=\textwidth]
    {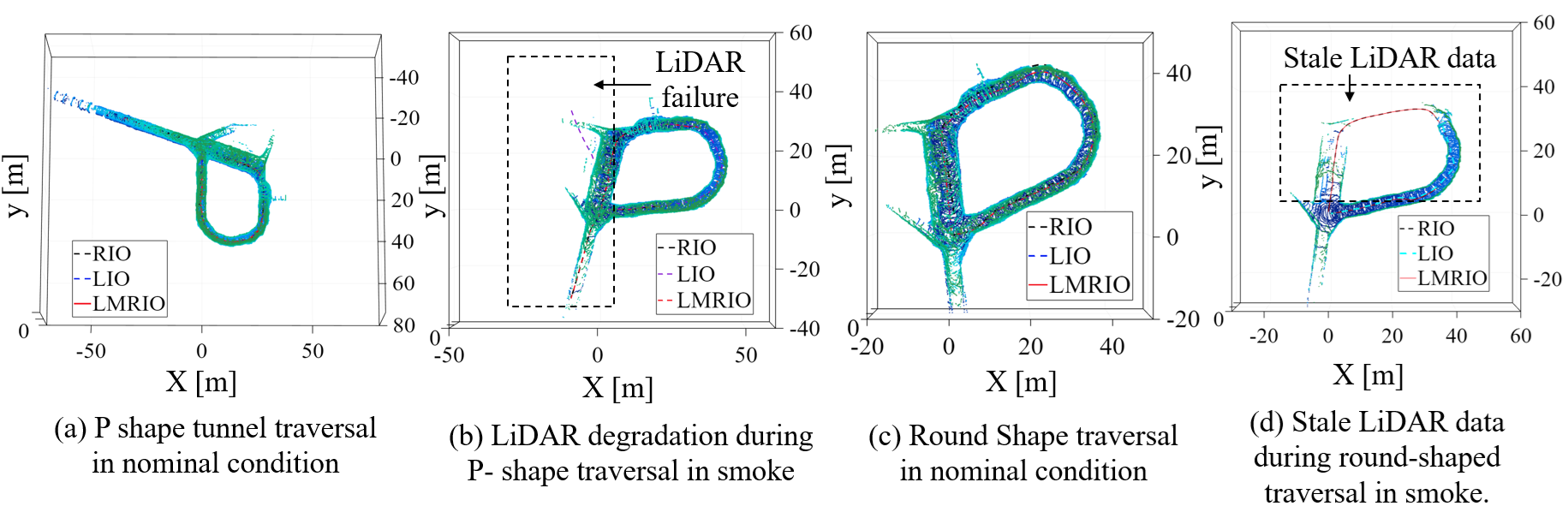}
    \caption{\justifying Localization trajectories and LiDAR maps for the P-shaped and round-tunnel traversals. Under nominal conditions, LMRIO closely follows the LIO estimate in both trajectories~(a,c). In dense smoke, the P-shaped traversal exhibits pronounced LIO divergence as LiDAR quality deteriorates~(b), while the round-tunnel traversal shows stale LiDAR measurements and loss of LiDAR-based localization updates~(d). In both degraded cases, LMRIO switches to RIO during intervals identified as unreliable, maintaining a consistent localization output.}
    \label{fig:localization_maps}
\end{figure*}
\begin{figure*}[h]
    \centering
    \includegraphics[width=0.99\textwidth]
    {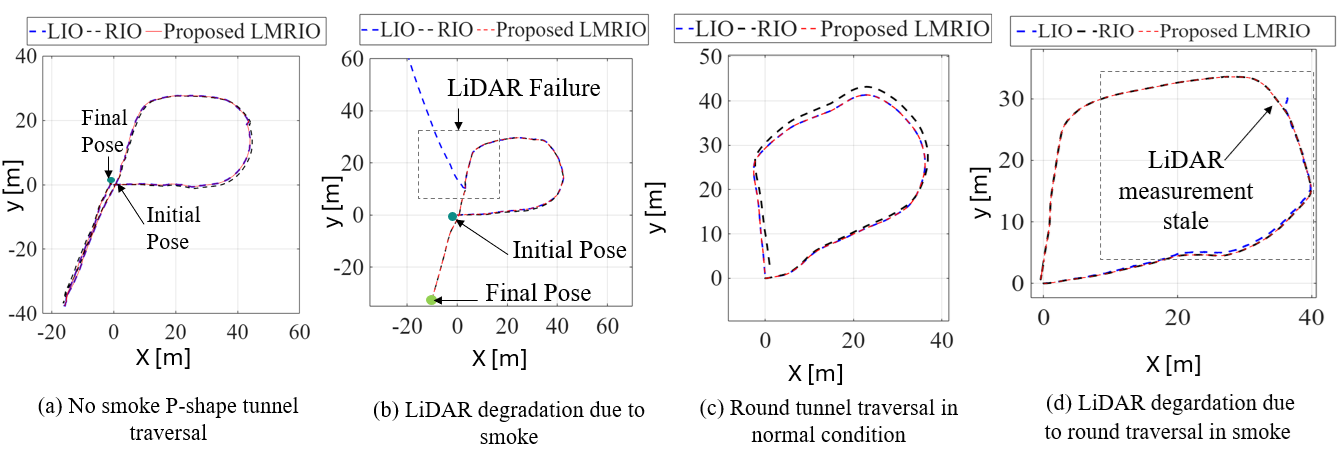}
    \caption{\justifying Trajectory comparison across the four experimental scenarios. Under nominal conditions, LMRIO closely follows LIO for both the P-shaped and round-tunnel traversals~(a,c). Under dense smoke, the P-shaped traversal exhibits severe LIO divergence following LiDAR degradation~(b), while the round-tunnel traversal shows stale LiDAR-based localization~(d). In both degraded cases, LMRIO selects RIO during unreliable LiDAR operation, preserving a continuous localization output.}
    \label{fig:trajectory_comparison}
\end{figure*}
\begin{figure*}[h]
    \centering
    \includegraphics[width=1.0\textwidth]
    {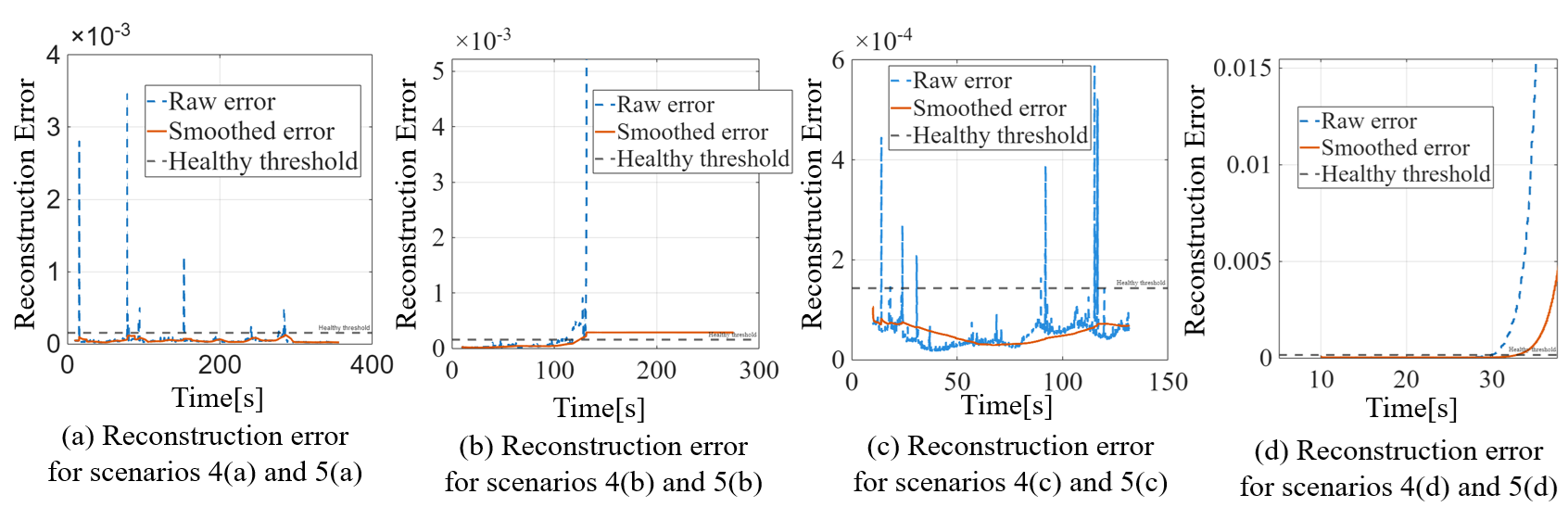}
    \caption{\justifying Raw and smoothed LSTM-autoencoder reconstruction errors with the nominal threshold. Nominal traversals remain largely below the threshold despite transient fluctuations~(a,c), whereas dense-smoke traversals show pronounced or sustained error growth associated with LiDAR degradation~(b,d). The smoothed error drives learned degradation detection, while LiDAR recency is evaluated independently.}
    \label{fig:reconstruction_results}
\end{figure*}
\begin{figure*}[h]
    \centering
    \includegraphics[width=0.99\textwidth]
     {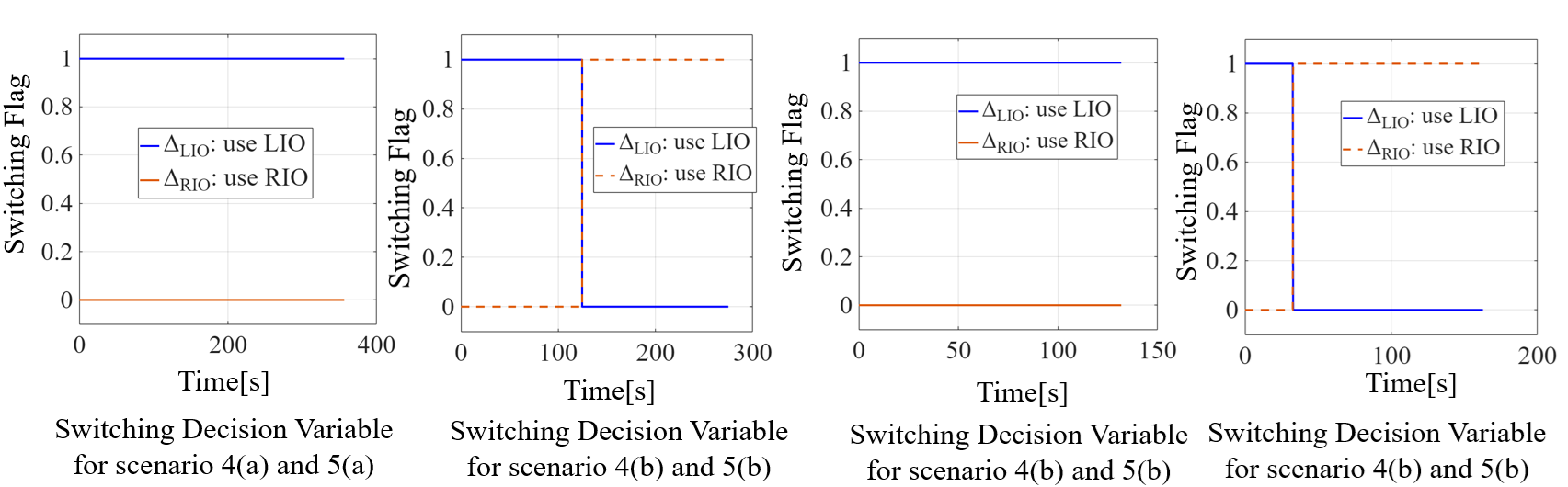}
    \caption{\justifying Estimator-selection traces across the four experimental scenarios. LIO remains selected throughout nominal operation~(a,c). Under dense smoke, the degradation detector triggers a transition to RIO in the P-shaped traversal~(b), while the LiDAR-recency criterion activates RIO when measurements become stale in the round-tunnel traversal~(d). The switching variables $\Delta_{\mathrm{LIO}}$ and $\Delta_{\mathrm{RIO}}$ indicate the active localization modality.}
    \label{fig:switching_decision}
\end{figure*}

\subsection{Localization Performance Under Nominal and Dense-Smoke Conditions}
\label{sec:nominal_smoke_results}

The nominal traversals are investigated with the proposed framework(Algo.~\ref {alg:lmrio}) to assess whether the reliability mechanism preserves LiDAR based localization during healthy sensing. For both the P-shaped and round-tunnel trajectories, LMRIO remains closely aligned with LIO
(Figs.~\ref{fig:localization_maps} and~\ref{fig:trajectory_comparison}), with no sustained degradation detected or unnecessary transition to RIO. The dense smoke experiments expose two complementary failure modes. In the P-shaped traversal, LiDAR measurements remain available but progressively degrade, leading to pronounced LIO divergence. In the round-tunnel traversal, LiDAR updates cease, and the LiDAR based estimate becomes stale. LMRIO responds to both cases by selecting RIO during degraded intervals, maintaining localization despite either measurement corruption or complete measurement staleness. These experiments therefore evaluate the framework against both LiDAR quality degradation and measurement unavailability.
\begin{table}[h]
\footnotesize
\centering
\caption{Configuration Parameters for 3D-Velodyne LiDAR}
\setlength{\tabcolsep}{2.25mm}
\renewcommand{\arraystretch}{1.2} %
\begin{tabular}{|c|c|c|c|c|}
\hline
\centering
\textbf{Parameter} & \textbf{Value}\\ \hline
\centering
\textbf{Laser Channels} &16  \\ \hline
\textbf{Vertical Resolution} &0.2° \\ \hline
\textbf{Field of View (FOV)} & 360°  \\ \hline
\textbf{Range} & 100m  \\ \hline
\textbf{Rotation Rate} &(1-100) Hz\\ \hline
\textbf{Points per second} &~300,000 points/sec  \\ \hline
\end{tabular}
\label{Table: 3D velodyne Lidar}
\end{table}
\begin{table}[h]
\footnotesize
\centering
\caption{Configuration Parameters for the TI-IWR6843AoP Millimeter-Wave Radar}
\label{Table:radar_specifications}
\setlength{\tabcolsep}{2.25mm}
\renewcommand{\arraystretch}{1.2}
\begin{tabular}{|l|c|}
\hline
\textbf{Parameter} & \textbf{Value} \\ \hline
Operating frequency range & 60--64 GHz \\ \hline
Available RF bandwidth & up to 4 GHz \\ \hline
Maximum IF bandwidth & 10 MHz \\ \hline
ADC sampling rate & up to 25 Msps \\ \hline
Number of transmitters & 3 \\ \hline
Number of receivers & 4 \\ \hline
\end{tabular}
\end{table}
%
\begin{table*}[h]
\centering
\caption{Ablation results for the P-shaped dense-smoke traversal.  The reconstruction-error-based switching time closest to the age criterion staleness time is highlighted.}
\label{tab:ablation_results}
\footnotesize
\setlength{\tabcolsep}{5pt}
\renewcommand{\arraystretch}{1.10}
\begin{tabular}{@{}lccccccc@{}}
\hline
\textbf{Variant} &
\shortstack{\textbf{LIO usage}\\\textbf{[\%]}} &
\shortstack{\textbf{RIO usage}\\\textbf{[\%]}} &
\shortstack{\textbf{First switch}\\\textbf{[s]}}&
\shortstack{\textbf{LiDAR Stale}\\\textbf{[s]}} &
\shortstack{$\boldsymbol{\Delta t_{\mathrm{stale}}}$\\\textbf{[s]}} &
\shortstack{\textbf{LMRIO-LIO}\\\textbf{RMSE [m]}} &
\shortstack{\textbf{LMRIO-RIO}\\\textbf{RMSE [m]}} \\
\hline
$C_1$ & 28.21 & 71.80 & 78.00  &130& 52.00 & $1.816\times10^{4}$ & 1.1148 \\
$C_2$ & 45.30 & 54.70 & \textbf{125.00} &130& 5.00  & $1.816\times10^{4}$ & 1.9943 \\
$C_3$ & 28.47 & 71.54 & 78.27  &130& 51.73 & $1.816\times10^{4}$ & 1.1287 \\
$C_4$ & 27.07 & 72.93 & 74.44  &130& 55.56 & $1.816\times10^{4}$ & 1.0553 \\
$C_5$ & 28.11 & 71.89 & 80.12  &130 &49.88 & $1.816\times10^{4}$ & 2.1200 \\
$C_6$ & 27.88 & 72.12 & 76.66 & 130 & 53.34 &
$1.816\times10^{4}$ & 1.0977 \\
\hline
\end{tabular}
\end{table*}
\subsection{Temporal Degradation Detection and Adaptive Modality Selection}
\label{sec:detection_switching_results}

The temporal detector identifies LiDAR degradation from persistent departures from nominal feature dynamics. As shown in Fig.~\ref{fig:reconstruction_results}, the raw reconstruction error contains short-term fluctuations during healthy operation, while moving average smoothing suppresses isolated peaks and preserves sustained deviations. In the nominal traversals, the smoothed score remains predominantly below the healthy threshold, and LIO is retained. In the P-shaped dense smoke traversal, observed in Fig.~\ref{fig:trajectory_comparison}, persistent reconstruction-error growth indicates progressive LiDAR degradation and triggers a transition to RIO before complete measurement loss. In the round-tunnel smoke traversal scenario, LiDAR updates eventually cease, preventing further reconstruction-error updates.

To handle this failure mode, LiDAR age is evaluated independently on the IMU timeline using a $1.5$~s staleness threshold. The final degradation decision is activated when either the learned anomaly criterion or the recency criterion is satisfied. Consequently, the learned detector handles degraded but available LiDAR measurements, while the age criterion captures missing or stale updates. Figure~\ref{fig:switching_decision} shows the resulting LIO-RIO selection across all four traversals.

\subsection{Feature Ablation Study}
\label{sec:ablation_results}

The feature family ablation study evaluates the influence of individual reliability cues on degradation timing, estimator usage, and trajectory agreement. Table~\ref{tab:feature_configurations} defines six variants, $C_1$-$C_6$. Configuration $C_1$ uses the complete twelve-feature representation, while $C_2$-$C_5$ remove the spatial, cross-modal, geometry/motion, and intensity groups, respectively. Configuration $C_6$ is a decision-level ablation in which the LiDAR-age feature remains an autoencoder input, but the independent recency criterion is disabled. Feature-level variants are retrained using the same nominal P-shaped data, whereas $C_6$ reuses the full $C_1$ model and modifies only the final
decision rule.

Table~\ref{tab:ablation_results} summarizes the P-shaped dense-smoke results. The first LIO to RIO transition ranges from 74.44 to 125.00~s, with RIO usage between 54.70\% and 72.93\%, demonstrating that feature composition directly affects switching behaviour. Configuration $C_2$, which removes the spatial statistics, switches latest at 125.00~s, approximately 5~s before LiDAR staleness at 130.00~s. This indicates that the remaining cues can still detect degradation before complete measurement
loss. A similar response is observed in the round-tunnel traversal, where switching occurs at approximately 35~s and LiDAR staleness at approximately 90~s.

The pre-staleness interval is defined as
\begin{equation}
\Delta t_{\mathrm{stale}}
=
t_{\mathrm{stale}}-t_{\mathrm{sw}},
\label{eq:pre_staleness_interval}
\end{equation}
where $t_{\mathrm{sw}}$ is the first switching time and $t_{\mathrm{stale}}=130.00$~s for the P-shaped dense-smoke traversal. Across configurations, the LMRIO-RIO RMSE ranges from 1.0553 to
2.1200~m, with the lowest value obtained by $C_4$. These results reveal a trade-off between degradation-response timing, estimator usage, and trajectory agreement, while showing that the full twelve-feature
representation is not necessarily optimal across all evaluated metrics. The large LMRIO-LIO RMSE under dense smoke is attributed to severe LIO divergence rather than ordinary localization uncertainty.
\begin{table}[h]
\centering
\caption{Feature-family configurations used in the ablation study.}
\label{tab:feature_configurations}
\footnotesize
\setlength{\tabcolsep}{2.5pt}
\renewcommand{\arraystretch}{1.08}

\begin{tabular}{@{}p{0.29\columnwidth}p{0.60\columnwidth}c@{}}
\hline
\textbf{Variant} & \textbf{Removed cues} & \textbf{Dim.} \\
\hline

Twelve Feature($C_1$)
& None
& 12 \\

No Spatial($C_2$)
& $N_\eta^{\mathcal L},\;
   N_\eta^v,\;
   \mu_\eta^d,\;
   \tilde{d}_\eta,\;
   \sigma_\eta^d,\;
   \mathrm{IQR}_\eta^d$
& 6 \\

No Cross-modal($C_3$)
& $\Delta p_{k_\eta}$
& 11 \\

No Geom./Motion($C_4$)
& $\Delta c_\eta,\;
   \omega_{k_\eta}^{I}$
& 10 \\

No Intensity ($C_5$)
& $\tilde{I}_\eta,\;
   \mathrm{IQR}_\eta^I$
& 10 \\

No LiDAR age ($C_6$)$^{\ast}$
& Explicit LiDAR-age decision rule
& 12 \\

\hline
\end{tabular}

\par\smallskip
\begin{minipage}{\columnwidth}
\footnotesize
$^{\ast}$The LiDAR-age feature $\tau_{k_\eta}^{\mathcal L}$ remains an
autoencoder input; only the independent recency criterion in the final
degradation decision is disabled.
\end{minipage}
\end{table}
\subsection{Quantitative Analysis}
\label{sec:quantitative_results}

Table~\ref{tab:trajectory_consistency} reports pairwise planar position RMSE for the P-shaped traversal. Because independent external ground truth is unavailable, these values measure agreement between localization estimates rather than absolute trajectory accuracy. Under nominal conditions, the LMRIO-LIO RMSE is 0.0121~m, compared with 0.0650~m between RIO and LIO. The small LMRIO-LIO discrepancy is consistent with the estimator-selection results, which show that LIO is retained throughout healthy operation. Under dense smoke, the LMRIO-RIO RMSE is 1.1148~m, whereas the LIO-LMRIO RMSE increases to $1.816\times10^{4}$~m as the LiDAR-based solution diverges. Together with the reconstruction-error and switching traces, these pairwise comparisons show that the executed LMRIO trajectory remains closely associated with LIO when LiDAR sensing is reliable and shifts toward the radar-inertial estimate once degradation is detected. The quantitative results therefore complement the qualitative trajectory plots by characterizing how the adaptive estimator-selection mechanism changes the resulting localization output across nominal and degraded
sensing conditions.
\begin{table}[h]
\centering
\caption{Pairwise RMSE Between Localization Trajectories in the P-Shaped Tunnel}
\label{tab:trajectory_consistency}
\footnotesize
\setlength{\tabcolsep}{4pt}
\renewcommand{\arraystretch}{1.08}
\begin{tabular}{@{}llc@{}}
\hline
\textbf{Condition} & \textbf{Trajectory pair} & \textbf{RMSE [m]} \\
\hline
Nominal & LMRIO-LIO & 0.0121 \\
Nominal & RIO-LIO & 0.0650 \\
Dense smoke & LMRIO-RIO & 1.1148 \\
Dense smoke & LIO-LMRIO & $1.816\times10^{4}$ \\
\hline
\end{tabular}
\end{table}
\section{Conclusions}

This article presented a one-class, reliability-aware LMRIO framework for resilient localization in GNSS-denied subterranean environments affected by dense smoke. LiDAR reliability is characterized using complementary spatial, temporal, cross-modal, geometric, inertial, and intensity cues, whose temporal evolution is modelled by an LSTM autoencoder trained exclusively on nominal data. The learned reconstruction error criterion is combined with an independent measurement recency check to address both progressive LiDAR degradation and complete measurement staleness. The resulting reliability decision enables online selection between LIO and RIO, retaining LiDAR-based localization during reliable sensing and transferring localization authority to radar-inertial odometry when LiDAR becomes degraded or unavailable. Real-world tunnel experiments under nominal and dense-smoke conditions demonstrated effective degradation detection, adaptive estimator switching, and continued localization output under both corrupted and stale LiDAR measurements. The feature-family ablation study further showed that a reduced representation can remain effective, and in some evaluated configurations improve switching behaviour and trajectory agreement, highlighting the importance of feature parsimony in reliability estimation. Future work will investigate generalization across different environments, obscurant types and alternative one-class degradation-detection methods, together with evaluation using independent trajectory ground truth where available.
\bibliographystyle{IEEEtran}
\bibliography{bibliography.bib}
\end{document}